\documentclass{article}

\usepackage{arxiv}

\usepackage[utf8]{inputenc} 
\usepackage[T1]{fontenc}    
\usepackage{hyperref}       
\usepackage{url}            
\usepackage{booktabs}       
\usepackage{amsfonts}       
\usepackage{nicefrac}       
\usepackage{microtype}      
\usepackage{graphicx}      
\usepackage{lipsum}
\usepackage{makecell, tabularx}
\title{SkyCam: \\ A Dataset of Sky Images and their Irradiance values}

\author{
  Evangelos Ntavelis \\
  Robotics \& Deep Learning, CSEM SA\\
  Computer Vision Lab, ETH Zurich \\
  Alpnach, Switzerland \\
  \texttt{evangelos.ntavelis@csem.ch} \\
   \And
   
 Jan Remund \\
  Energy \& Climate\\
  Meteotest \\
  Bern, Switzerland \\
  \texttt{jan.remund@meteotest.ch} \\   
  
  \And
  
 Philipp Schmid \\
  Industry 4.0 \& Machine Learning \\
  CSEM SA\\
  Alpnach, Switzerland \\
  \texttt{philipp.schmid@csem.ch} \\
}

\begin{document}
\maketitle

\begin{abstract}

Recent advances in Computer Vision and Deep Learning have enabled astonishing results in a variety of fields and applications. Motivated by this success, the SkyCam Dataset aims to enable image-based Deep Learning solutions for short-term, precise prediction of solar radiation on a local level. For the span of a year, three different cameras in three topographically different locations in Switzerland are acquiring images of the sky every 10 seconds. Thirteen high resolution images with different exposure times are captured and used to create an additional HDR image. The images are paired with highly precise irradiance values gathered from a high-accuracy pyranometer.  

\end{abstract}


\section{Introduction}

Recent years have seen the rise of Photovoltaic (PV) energy as a solution to the problem of sustainable energy. Moreover, a move to local producers of energy is observed compared to the traditional centralized production systems \cite{photovolt}. Thus, it is important to be able to predict the energy yield of a PV system especially in stand-alone solutions.

Stand-alone PV systems operate independently of a central electric grid. When a PV system is used in conjunction with an engine-generator or utility power it is referred as a photovoltaic-hybrid-system\cite{BHATIA2014144}. In such cases it is important to be able to forecast with high accuracy the solar radiance and thus the energy output of the PV system in order to predict the need to switch to the alternative energy generator.

The motivation behind the preparation of the SkyCam dataset is to gauge the effectiveness of a low-cost and camera-based solution on predicting the solar irradiance values on a local scale\cite{skycam}. This information would potentially improve the energy-performance of stand-alone solutions, where PV panels are used in conjunction with diesel generators to generate power.  Since the generator would not have to run permanently on standby, a more accurate forecast could greatly reduce the consumption of fossil fuels \cite{skycam}. 

The dataset offers:
\begin{itemize}
    \item Data from 3 topographically diverse locations
    \item Collection of images over the span of a whole year
    \item A sample every 10 seconds
    \item 13 images of different exposure times and a High-Dynamic-Range post-process image
    \item Precise solar radiance measurements
\end{itemize}

\begin{table}
 \caption{Specifications table}
 \small
 \setlength\tabcolsep{3pt}
\centering
  \begin{tabularx}{\linewidth}{@{} ll}
    \toprule
    Subject area    &   Computer Vision, Renewable Energy, Meteorology \\
    \addlinespace
    
    More specific subject area    &  Irradiance prediction, Cloud movement prediction, Weather prediction \\
    \addlinespace
    Type of data    & 13 concurrent images with different exposure times, \\
    & 1 HDR post-processed image, 1 irradiance value per 10 secs \\
    \addlinespace
    How  data was acquired & 1 camera sensor and 1 pyranometer pointing to the sky, at each of the three different locations \\
    \addlinespace
    Data format    &   Png and XML files \\
    \addlinespace
    Data source location    &   The images were acquired by cameras on three locations across Switzerland: \\
    & CSEN Alpnach, University Hospital Bern, CSEM Neuchâtel \\
    \addlinespace
    Data accessibility    &   The data is available for download at: \url{https://github.com/vglsd/SkyCam} \\
    \bottomrule
  \end{tabularx}
  \label{tab:specifications}
\end{table}

\section{Collection of the data}
\label{sec:dataset}

Switzerland is a perfect testing ground for creating a dataset that aims to tackle the problem of short-term solar radiation prediction due to its diverse topographical features and highly variable meteorological conditions\cite{brunner}. To this extend image and solar radiance capturing equipment was set to three topographically distinct locations, acquiring observations over the span of a year.

\subsection{Camera Locations}

\begin{figure*}[h]
    \centering%
    \begin{tabular}{cc}
         \includegraphics[width=0.37\linewidth]{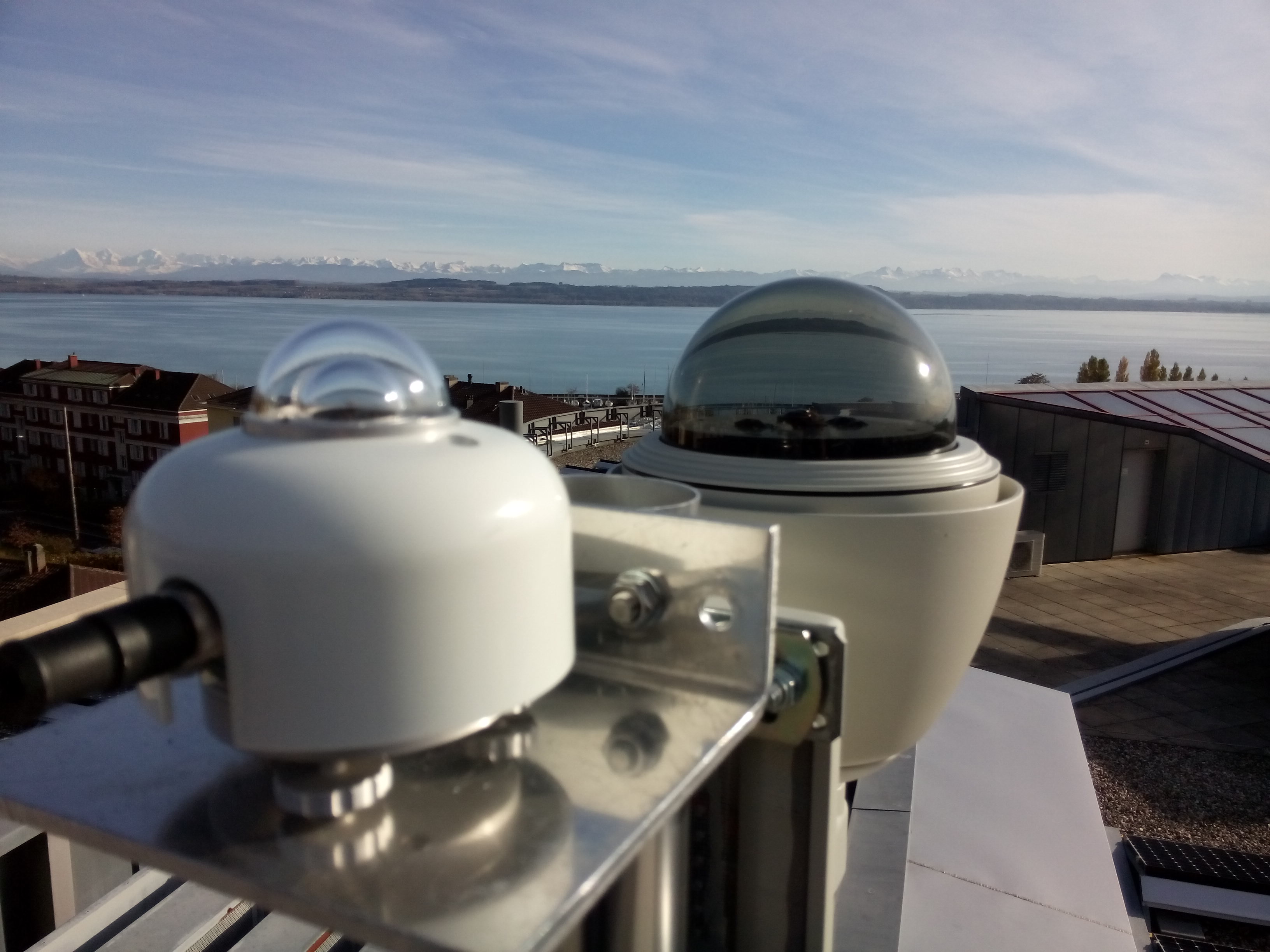} &
         \includegraphics[width=.58\linewidth]{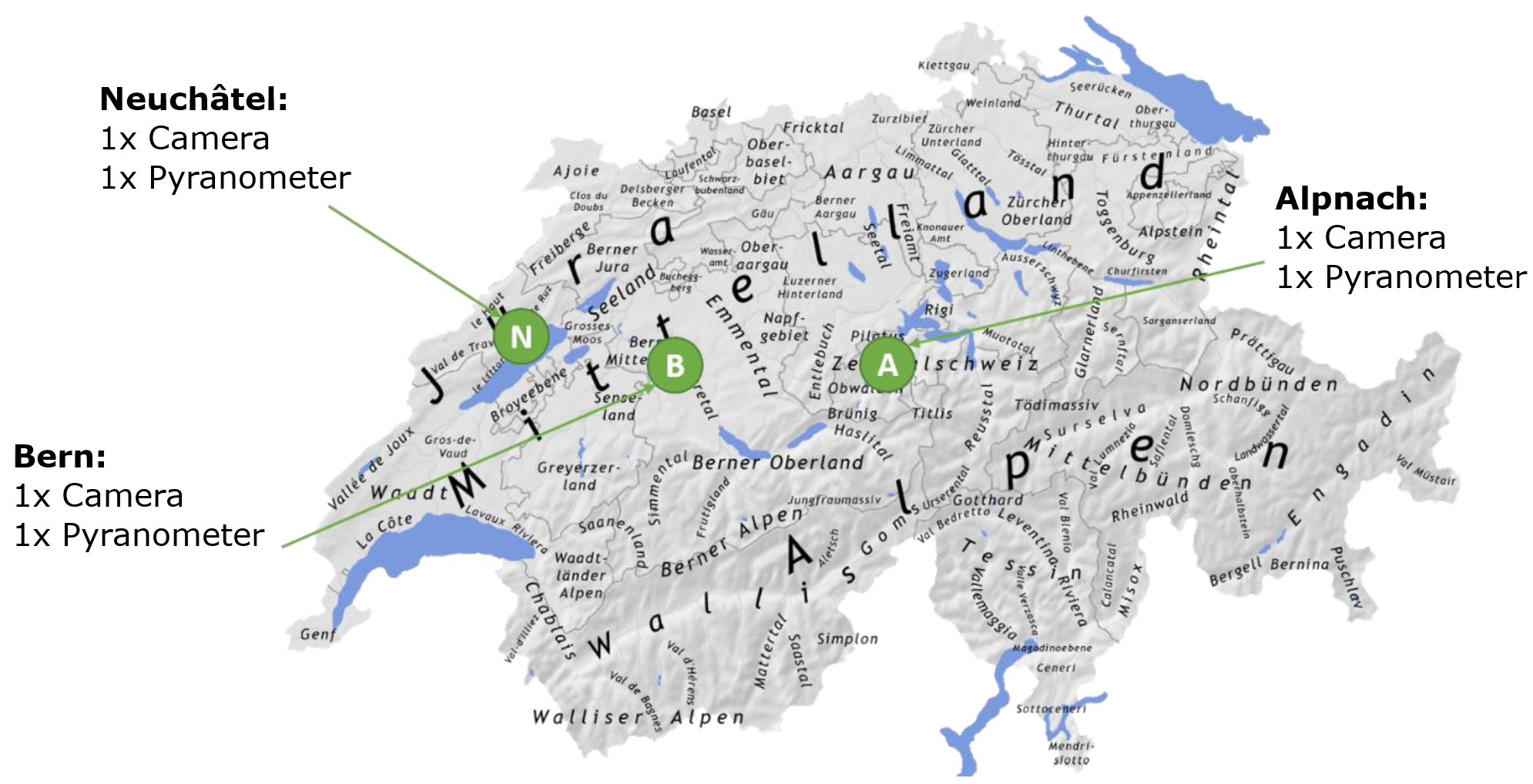} \\
        (a) & (b) \\
    \end{tabular}
    \caption{\textbf{(a)} The equipment used for the data acquisition: a high accuracy pyranometer(left) and a fish-eye camera (right) \textbf{(b)} The image capturing sites. The three different locations across Switzerland are characterized by different topological conditions. }
    \label{fig:cammap}
\end{figure*}

The variable topography of Switzerland results in diverse hydro-meteorological conditions \cite{brunner} and the locations where the images are acquired are diverse and characterized by a unique topology compared to one another. Neuchâtel is in the region of Jura, Bern is the Swiss Plateau and Alpnach is in the Pre-Alps area (Fig. \ref{fig:cammap}.b).  Their first difference is in elevation: Alps and Jura regions has high elevations but the Swiss Plateau has low \cite{brunner}.

\subsection{Collection Times}

\begin{figure*}[h]
\centering
\includegraphics[width=.8\linewidth]{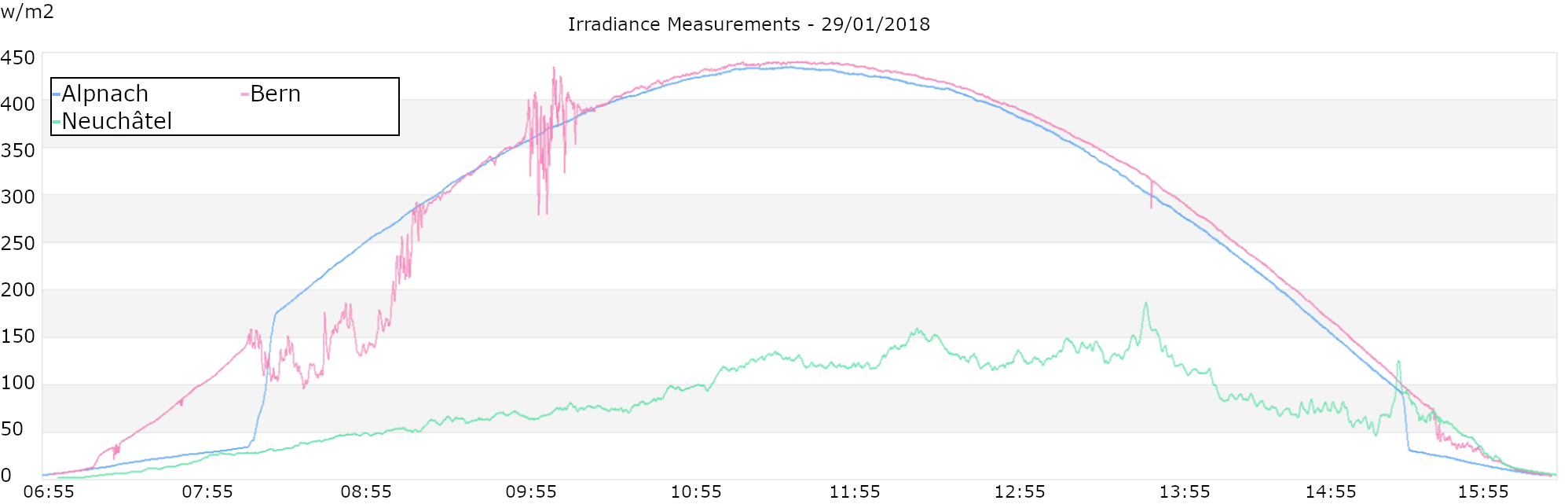}
\caption{Irradiance measurements at the three different locations on the same day. A lag of approximately 3 minutes can be observed between the Bern and Alpnach sites. As it is shown in the graph the three different locations can have vastly different weather conditions with one having a very cloudy sky while the others a clear one. This results in great difference in the levels of the irradiance values}
\label{fig:lag}
\end{figure*}

The data was collected across the span of the year 2018. Consequently, we aim to capture the diverse seasonal effects created by the topographical differences between the collections sites. On each day of the year the equipment starts acquiring pictures and measurements at the time of dawn and stops at dusk. As the three locations have different longitudes and thus different times for dawn, peak sun point and dusk. This can be observed in Figure \ref{fig:lag}. The samples were collected with a frequency of 10 seconds.  

\subsection{Equipment}

In all three locations a single industrial camera was used. In all sites a high-accuracy pyranometer was deployed in order to acquire the irradiance values with high precision. All cameras are industrial-grade CMOS cameras with an appropriate protective casing. The pyranometer is a Hukseflux SR30 designed for for Photovoltaic system performance monitoring and meteorological networks. At the start of the measurements the pyranometer has been calibrated. The solar radiation is received by a horizontal surface and measured in $W/m^2$, from a 180{\textdegree} field of view angle\cite{hukseflux, huksefluxmemo}.

\section{SkyCam Dataset}

\begin{figure*}[h]
    \centering
    \setlength{\tabcolsep}{-.25pt}
    \begin{tabular}{ccccc}
         \includegraphics[width=0.185\linewidth]{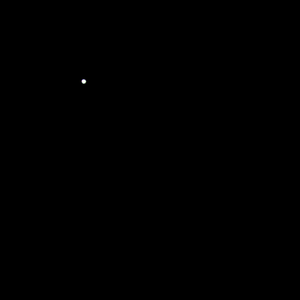} &
         \includegraphics[width=0.185\linewidth]{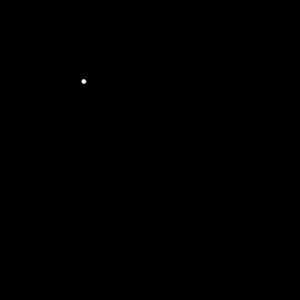} &
         \includegraphics[width=0.185\linewidth]{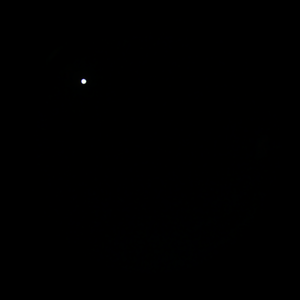} &
         \includegraphics[width=0.185\linewidth]{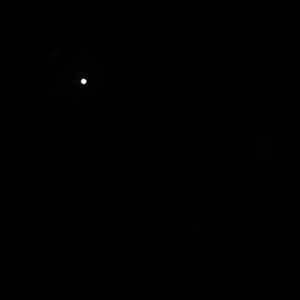} &
         \includegraphics[width=0.185\linewidth]{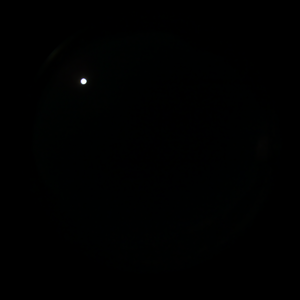} \\
         \includegraphics[width=0.185\linewidth]{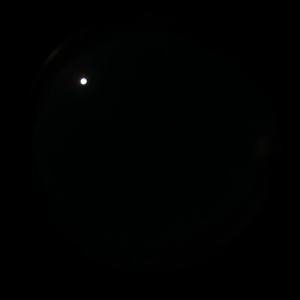} &
         \includegraphics[width=0.185\linewidth]{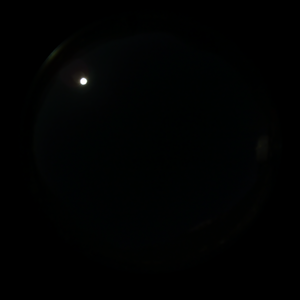} &
         \includegraphics[width=0.185\linewidth]{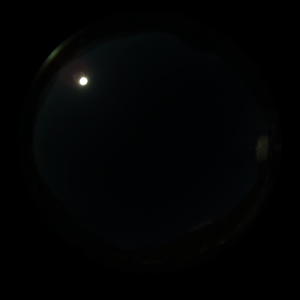} &
         \includegraphics[width=0.185\linewidth]{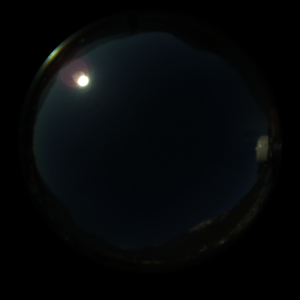} &
         \includegraphics[width=0.185\linewidth]{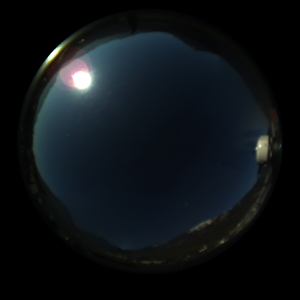} \\
         \includegraphics[width=0.185\linewidth]{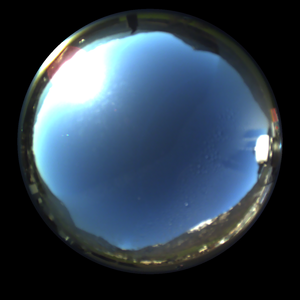} &
         \includegraphics[width=0.185\linewidth]{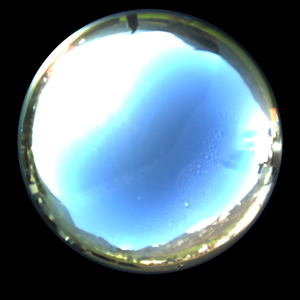} &
         \includegraphics[width=0.185\linewidth]{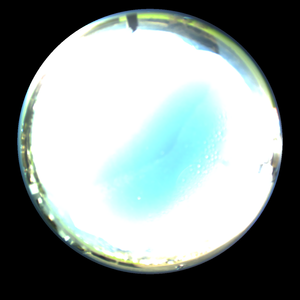} &
         &
         \includegraphics[width=0.185\linewidth]{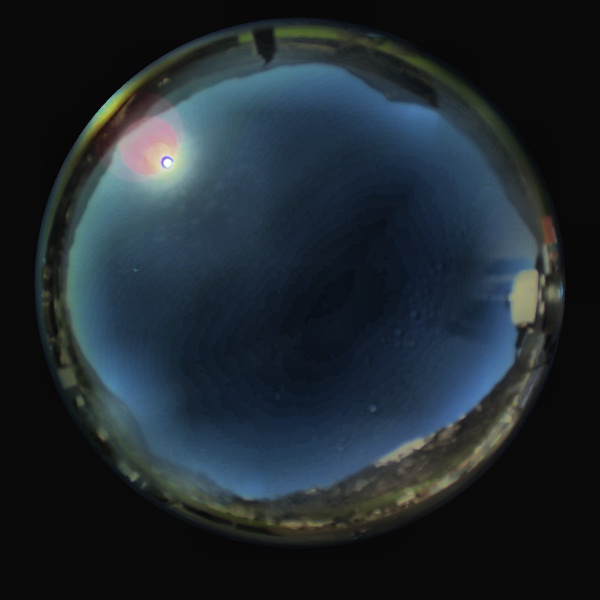} \\
    \end{tabular}
    \caption{All images produced by the system on the 29th of January 2018 at 11:09:20 on Alpnach site. The 13 first images were captured sequentially with increasing exposure times. The last is the HDR image produced post-capture. The irradiance value of the timestamp is 452.94 $W/m^2$ }
    \label{fig:example}
\end{figure*}

The SkyCam dataset is a collection of images from 365 days from three different locations and three cameras.
Each day has on average 12 hours between dawn and dusk and images are captures with a frequency of 10 seconds. That means that we have approximately 1576800 timestamps per camera. The total size of the dataset is approximately 16 TeraBytes.

\subsection{Data description of each timestamp}

For each time stamp each camera acquires a series of 13 photographs with different exposure times as well. After the collection the images are post-processed in order to generate a High-Dynamic-Range (HDR) image. At the same time the solar irradiance value is measured by the pyranometer. The measurements were not quality proofed. A description of the 13 photographs along with the irradiance values are included in a XML file. In Figure \ref{fig:example} we can observe how a particular timestamp looks like. The original resolution of each captured image is $1200\times1200$ pixels while the generated HDR image has a resolution of $600\times600$ pixels.

\subsection{Missing and Erroneous Data}
\begin{figure*}[h]
    \centering
    \setlength{\tabcolsep}{-.25pt}
    \begin{tabular}{ccc}
         \includegraphics[width=0.3\linewidth]{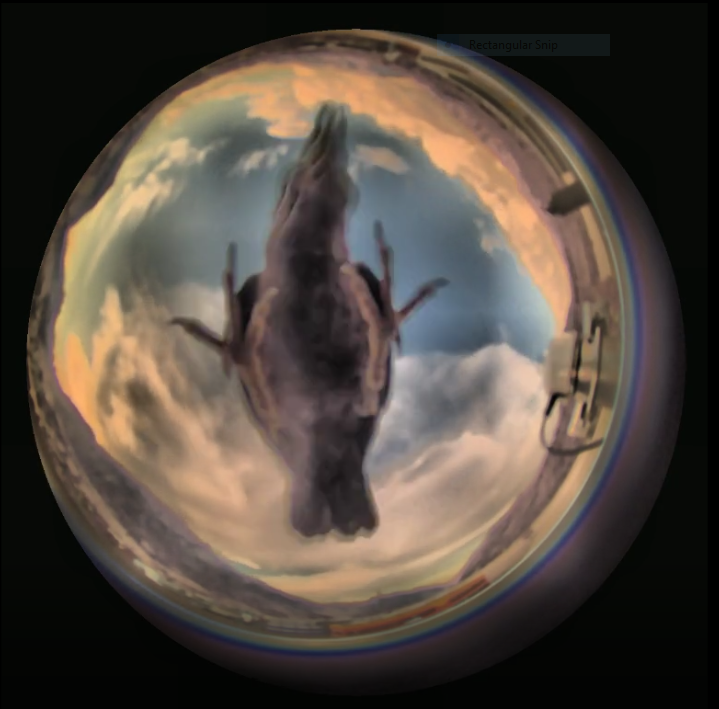} &
         \includegraphics[width=0.3\linewidth]{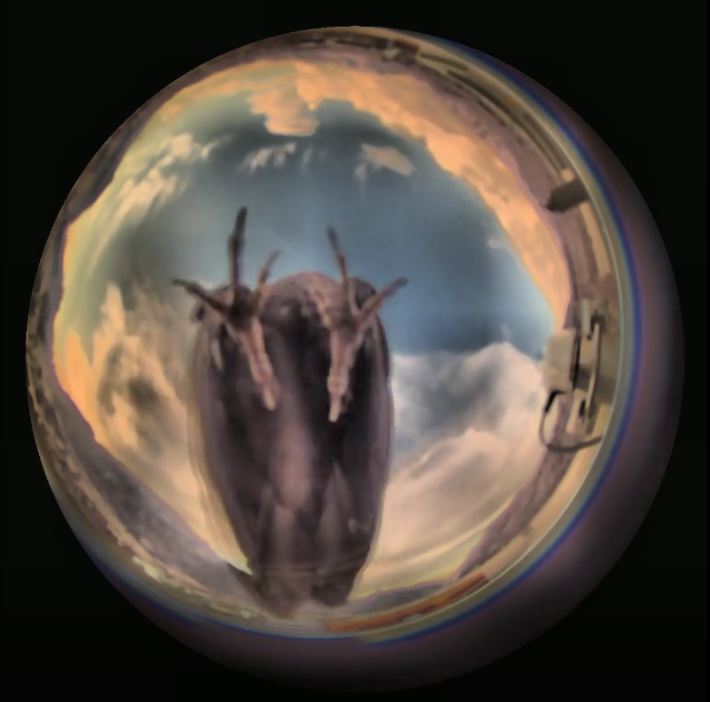} &
         \includegraphics[width=0.3\linewidth]{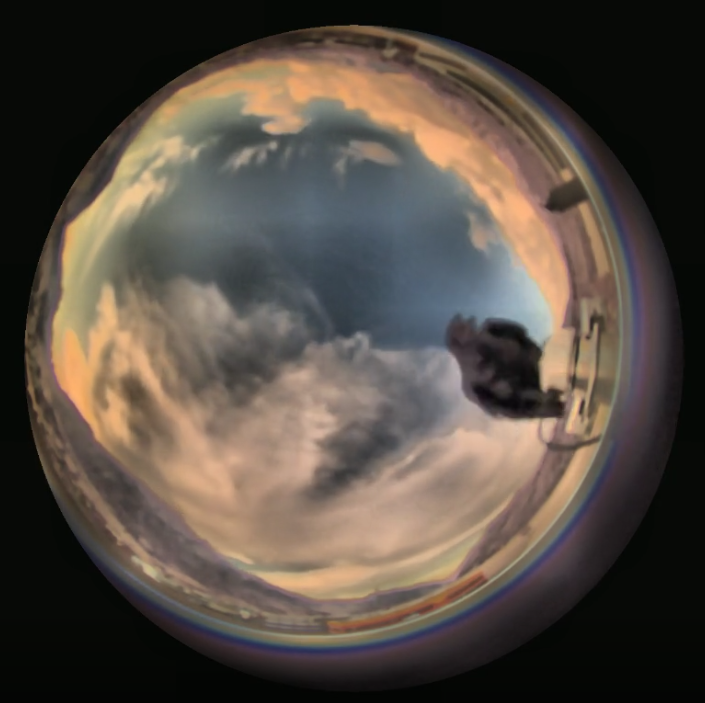} \\
    \end{tabular}
    \caption{Example cases of problematic timestamps where part of the sky is occluded }
    \label{fig:bird}
\end{figure*}

During the span of one year various problems lead to temporary disruptions of the sampling. These disruptions lead to the samples of certain timestamps to be missing in one of the locations. Moreover, in some cases, as seen in Figure \ref{fig:bird}, birds have occluded the view of the sky. These images are still in the dataset. In both cases this is left to the user to handle. 

\section{Conclusion}
In this article we present a description of the SkyCam dataset, its acquisition process and its characteristics. It offers a variety of locations with diverse topographical characteristics (Swiss Jura, Plateau and Pre-Alps regions), 
where a camera with a high-accuracy pyranometer are deployed.
The dataset is collected with a high frequency rate, a data sample is generated every 10 seconds. 13 images with different exposures times are generated along with a post-processed HDR images and a solar radiance values for each of the locations.
SkyCam dataset will enable future works to tackle the problem of short-term local camera-based solar radiance prediction. 

\section{Acknowledgements}
The  development  of  the  SkyCam dataset  was funded by the  Swiss  Federal  Office of  Energy  under  grant  SI/50151 as a collaborative project between CSEM and Meteotest.

\bibliographystyle{unsrt}  
\bibliography{refs}  

\end{document}